\documentclass[twoside,11pt]{article}

\usepackage{jmlr2e}
\usepackage{xcolor} 
\usepackage{float}
\usepackage{subfig}

\usepackage{amsmath,amsfonts,bm}
\usepackage{upgreek}

\def\eqref#1{equation~\ref{#1}}

\def\1{\bm{1}}

\def\ry{{\textnormal{y}}}

\def\vtheta{{\bm{\theta}}}

\def\vx{{\bm{x}}}

\DeclareMathAlphabet{\mathsfit}{\encodingdefault}{\sfdefault}{m}{sl}
\SetMathAlphabet{\mathsfit}{bold}{\encodingdefault}{\sfdefault}{bx}{n}

\ShortHeadings{Adaptive Experimental Design and Active Learning in the Real World}{Wang \& Nalisnick}
\firstpageno{1}

\begin{document}

\title{Active Learning for Multilingual Fingerspelling Corpora}

\author{\name Shuai Wang \email shuai.wang@student.uva.nl \\
       \name Eric Nalisnick \email e.t.nalisnick@uva.nl \\
       \addr Informatics Institute\\
       University of Amsterdam\\
       Amsterdam, Netherlands}


\maketitle

\begin{abstract}
We apply active learning to help with data scarcity problems in sign languages.  In particular, we perform a novel analysis of the effect of pre-training.  Since many sign languages are linguistic descendants of French sign language, they share hand configurations, which pre-training can hopefully exploit.  We test this hypothesis on American, Chinese, German, and Irish fingerspelling corpora.  We do observe a benefit from pre-training, but this may be due to visual rather than linguistic similarities. 
\end{abstract}


\section{Introduction}


	
While there has been recent advances in technologies for written languages (e.g.~BERT \citep{kenton2019bert}, GPT-3 \citep{brown2020language}), \emph{signed languages} have received little attention~\citep{yin-etal-2021-including}.\footnote{We thank \citet{desai-etal-2024-systemic} for pointing out that a sentence in a previous draft incorrectly implied that deaf people only benefit from technologies for signed languages.  It has been removed.}  One reason for this gap is that \textit{sign language processing} (SLP) is inherently more challenging since identifying signs is a complex computer vision task---compared to string matching for written languages.  Yet, the scarcity of sign language corpora is the largest hurdle to advances in SLP.  There are, of course, relatively fewer users of signed languages, leading to less available training data.  Moreover, the relative scarcity of native signers makes procuring high-quality labels and annotations for supervised learning difficult and costly.

\begin{figure}
    \centering
    \includegraphics[width=0.7\textwidth]{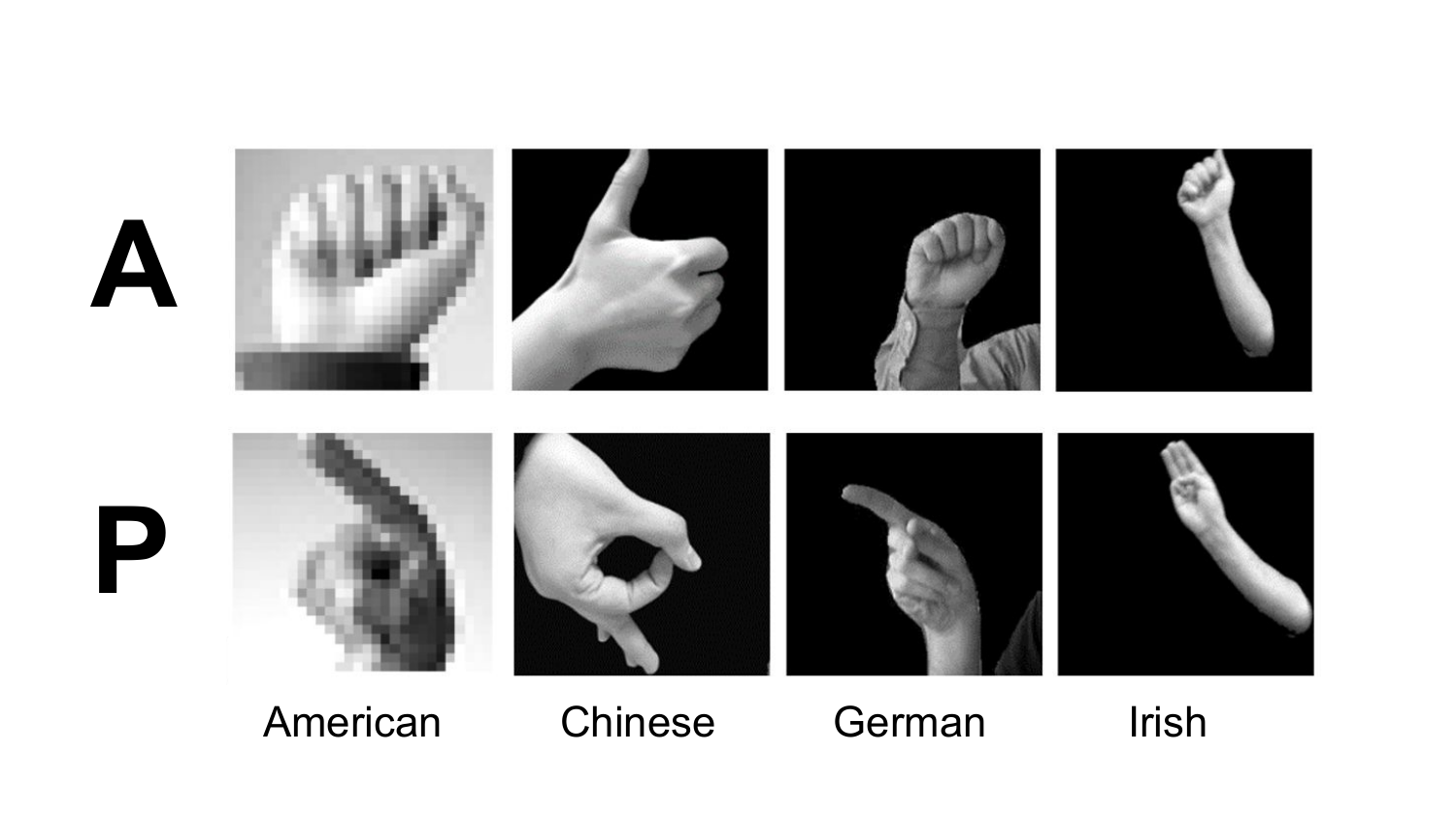}
    \vspace*{-7mm}
    \caption{\textit{Example Images}.  The above figure shows the letters `A' and `P' from the ASL, CSL, GSL, and ISL data sets respectively.}
    \vspace*{-5mm}
    \label{fig:dataset_sample}
\end{figure}

In this work, we make progress on alleviating data scarcity in SLP by applying \textit{active learning} (AL) \citep{settles2012active} to multilingual \textit{fingerspelling} corpora.  \textit{Fingerspelling} is a sub-task in sign language communication: the signer spells-out a word associated with a concept that does not have an associated sign, such as a proper noun.  For example, if a signer wants to reference ``GPT-3,'' the signer would make a hand shape for each letter `G,' `P', and `T'.  See Figure \ref{fig:dataset_sample} for example images.  In addition to detecting sign-less concepts, recognizing fingerspelling is important because the hand configuration for some concepts are the same as the first letter of the associated word.  For instance, in American sign language, the concept of `ready' is signed with the hands in the `r'-configuration.  

While fingerspelling recognition is a relatively easier task than full SLP, it can still be challenging for low-resource languages.  Consider Irish sign language: it has just $5000$ active deaf users, and to make matters more complicated, has variations depending on the gender of the signer.  Collecting large, diverse data sets for languages such as this one requires great effort.  Thus, we also perform \textit{transfer active learning} by exploiting linguistic relations across sign languages.  Specifically, American, Irish, German, and Chinese sign languages all are descendants of French sign language, thus resulting in shared hand configurations, among other features.  We conjecture that learning can be done more quickly by first pre-training a model on a related sign language and then performing AL on an especially low-resource target language. 

We report two sets of experiments.  In the first, we perform AL on American, Irish, German, and Chinese sign languages, demonstrating improvements over random sampling.  In the second, we perform transfer AL by first pre-training the model on one of the other three sign languages.  This is the first work to investigate leveraging linguistic relationships for transfer AL.  While we observe some benefits, our initial results suggest that pairing data sets with a similar visual style improves transfer AL more than having a strong linguistic relationship but dissimilar visual style.

\section{Data Sets: Multilingual Fingerspelling Corpora} \label{Datasets}
We use the following publicly available fingerspelling corpora, all of which have a linguistic relationship due to being descendants from French sign language.  In all cases, we standardize the image resolution to $28 \times 28$ pixels.  For each data set, we resample the frequency of each letter so that it follows the character distribution of the corresponding written language. As a result, the data sets are class imbalanced.
\begin{enumerate}
    \item \textbf{American Fingerspelling (ASL)}: This data set is availabe on Kaggle\footnote{\hyperlink{https://www.kaggle.com/datasets/datamunge/sign-language-mnist}{https://www.kaggle.com/datasets/datamunge/sign-language-mnist}} and contains $34,627$ grayscale images (resolution $28 \times 28$) for $24$ alphabetic characters used in American sign language (ASL).  The characters `J' and `Z' are omitted because they are dynamic signs that cannot be well-represented by a static image.  The images are closely cropped to the hand and in turn do not require pre-processing.
    \item \textbf{Chinese Fingerspelling (CSL)}~\citep{Jiang2020FingerspellingIF}: This data set consists of the alphabetic characters used in Chinese sign language (CSL).  Specifically, there are $1320$ images, each labeled as one of $30$ letters---$26$ are the characters A through Z and $4$ are double syllable letters (ZH, CH, SH, and NG). The resolution is originally $256 \times 256$, and we down-sample them to $28 \times 28$.  We discard the double syllable letters since they are unamenable to transfer learning.  We convert all images to grayscale and remove `J' and `Z' so that the data set is aligned with the ASL data set.
    \item \textbf{German Fingerspelling (GSL)}~\citep{dreuw06smvp}:  This data set contains alphanumeric characters used in German sign language (GSL).  The original data set consists of video sequences for $35$ gestures: the $26$ letters A to Z, the $4$ German umlauts (SCH, Ä, Ö, Ü), and the numbers from 1 to 5.  We extracted $20,904$ static images for the letters from the video frames.  We remove the background environment using a ResNet-101 model pre-trained for segmentation~\citep{segmentation}.  We again convert the images to grayscale and remove `J' and `Z'.
    \item \textbf{Irish Fingerspelling (ISL)}~\citep{inproceedings_Oliveira}: This data set contains the alphabetic characters used in Irish sign language (ISL).  It consists of $58,114$ images for the $26$ letters A through Z.  This data set is unique from the others in that the sign is repeated by moving the forearm from a vertical to near horizontal position.  Again, we convert the images to grayscale and remove `J' and `Z'.
\end{enumerate}




 Figure \ref{fig:dataset_sample} shows images for the letters `A' and `P' from the four corpora listed above.  The sign for `A' is very similar for ASL, GSL, and ISL.  The sign for `P' is the same for ASL and GSL but different for CSL and ISL. 


\section{Methodology: Active Learning}
Our aim is to obtain a predictive model for fingerspelling recognition.  Given an input image, the model should be able to predict a label representing the alphabetic character being signed.  Yet, given the resource constraints for signed languages (discussed in Section 1), we wish to train a highly accurate predictive model using as few labeled instances as possible.  To achieve this, we rely on the methodology of \textit{active learning} (AL) \citep{settles2012active}, which allows us to obtain a sequence of highly informative labels from an oracle (such as a human expert).  The hope is that the intelligent selection of these labels allows the model to achieve satisfactory performance while minimizing queries of the oracle.  Formally, we assume the model is first trained on an initial data set $\mathcal{D}_0 = \{\vx_n, y_n\}_{n=1}^N$, with $N$ being relatively small for the problem at hand.  We also assume access to an unlabeled pool set $\mathcal{X}_p = \{\vx_m\}_{m=1}^M$ such that $M >> N$.  The oracle has access to the corresponding labels for this pool set, denoted $\mathcal{Y}_p = \{y_m\}_{m=1}^M$.


AL for image data has been well-studied \citep{Bayesian_AL}, and various acquisition functions have been proposed. In preliminary experiments, we studied the following: maximum entropy \citep{shannon},
\textit{BALD} \citep{journals/corr/abs-1112-5745}, variation ratios ~\citep{Freeman1965ElementaryAS}, and mean standard deviation.  We found all methods to be comparable but with variation ratios to have a slight edge.  In turn, we use variation ratios in all experiments.  At time step $t$, a point from the pool set is selecting according to:
\begin{equation}
    \vx^\ast_{t} \ \  = \ \  \underset{\vx \in \mathcal{X}_{p,t-1}} {\text{arg max}} \quad 1 - \max_{\ry} p\left(\ry | \vx, \vtheta_{t-1}\right)
\end{equation} where $p\left(\ry | \vx, \vtheta_{t-1}\right)$ is the model's maximum class confidence after being fit during the previous time step. After obtaining $\vx^\ast_{t}$, the oracle is queried for the associated label, and the model is re-fit to data that includes the selected feature-label pair (to obtain $\vtheta_{t}$).  


\paragraph{Transfer Active Learning} Knowledge gained from other domains can be helpful to improve performance on the target domain.  In fact, \citet{pretrained_good_AL} suggest that the ability to actively learn is an emergent property of pre-training.  We combine this inspiration with the fact that ASL, CSL, GSL, and ISL are linguistically related, proposing to pre-train on a related fingerspelling corpora before performing AL on the target domain.  Specifically, we assume access to an auxiliary data set $\mathcal{D}_{-1} = \{\vx_l, y_l\}_{l=1}^L$.  We train the model of interest on this data set before training it on the pool set $\mathcal{D}_{0}$.  $\mathcal{D}_{-1}$ is seen by the model only once and not retained as part of the growing data set as AL proceeds.




\begin{figure}[ht]
  \subfloat[American (ASL)]{
	\begin{minipage}[c][1\width]{
	   0.24\textwidth}
	   \centering
	   \includegraphics[width=1\textwidth]{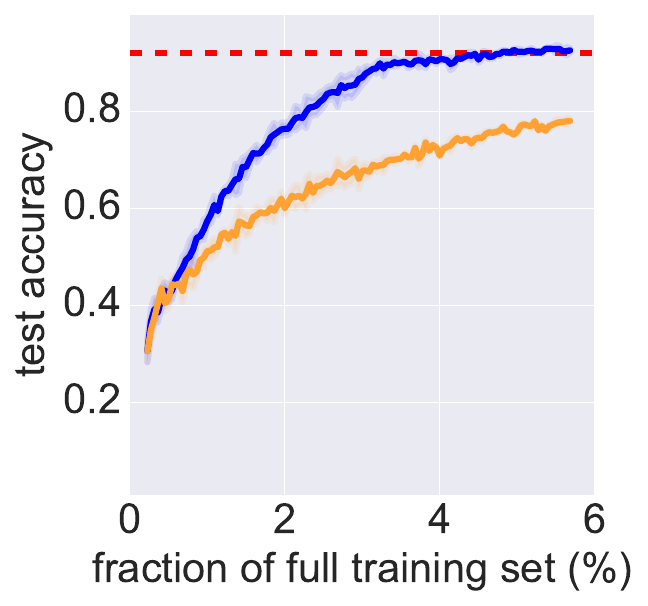}
	\end{minipage}}
 \hfill 	
  \subfloat[Chinese (CSL)]{
	\begin{minipage}[c][1\width]{
	   0.24\textwidth}
	   \centering
	   \includegraphics[width=1\textwidth]{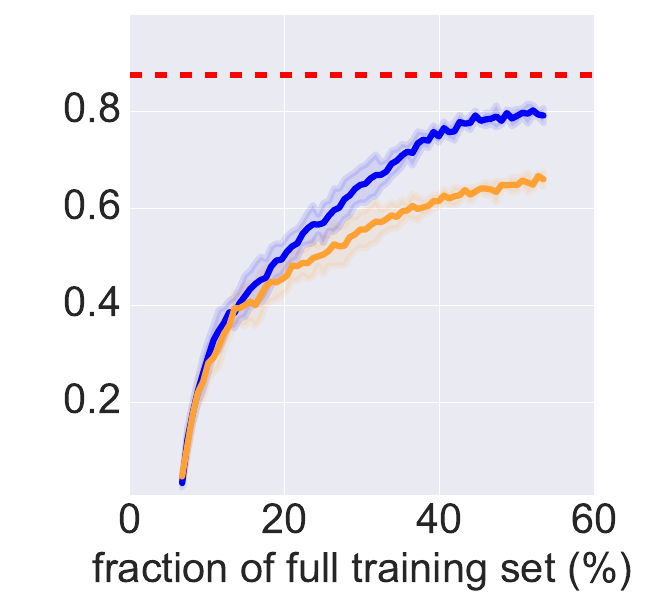}
	\end{minipage}}
 \hfill	
  \subfloat[German (GSL)]{
	\begin{minipage}[c][1\width]{
	   0.24\textwidth}
	   \centering
	   \includegraphics[width=1\textwidth]{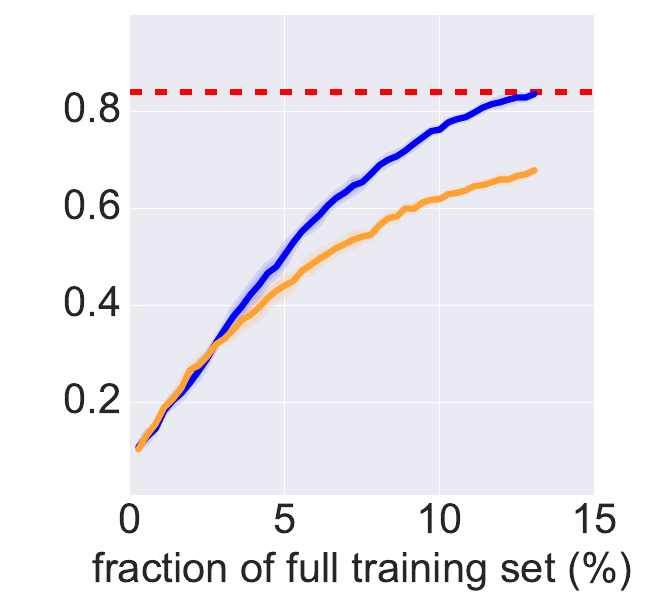}
	\end{minipage}}
 \hfill	
  \subfloat[Irish (ISL)]{
	\begin{minipage}[c][1\width]{
	   0.24\textwidth}
	   \centering
	   \includegraphics[width=1\textwidth]{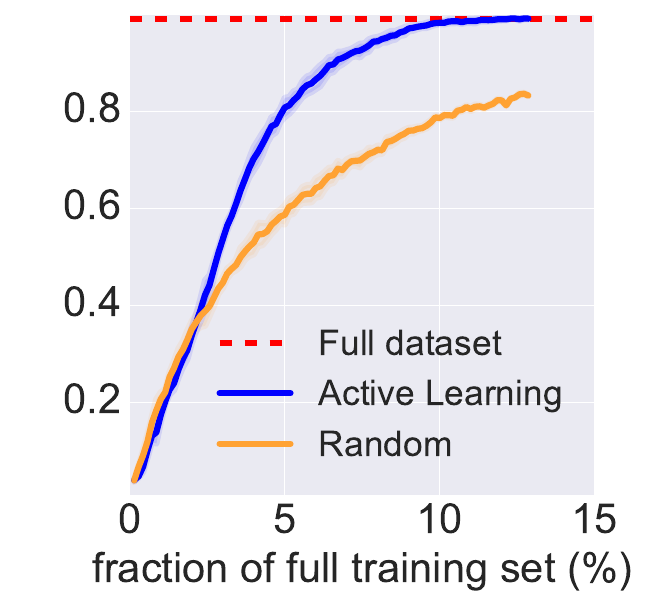}
	\end{minipage}}
\caption{\textit{Single Corpus Active Learning}: The figure above shows AL results for each fingerspelling corpus.  Using variation ratios (blue) as an acquisition function is clearly superior to random sampling (yellow).}
\label{fig:active_res}
\end{figure}

\section{Experiments}

We perform two types of experiment. For the first, we verify the effectiveness of AL for an individual fingerspelling corpus.  For the second, we examine if AL can be made more sample efficient by pre-training on a linguistically related corpus.  The following settings apply to all experiments.  We construct the initial set $\mathcal{D}_{0}$ by choosing two samples per class uniformly at random ($N = 48$). For constructing test sets, we hold-out $30\%$ of the data in CSL (because of its small size) and 10\% for the other corpora.  The remaining data is the pool set.  The query size for each data set is different due their different sizes: $50$ for ISL and GSL, $10$ for ASL, and $5$ for CSL.  The classifier is retrained from a random initialization whenever new labels are procured.  The classifier is trained for $50$ epochs using a batch size of $128$ with learning rate of $10^{-3}$.  


\begin{figure}[htbp]
    \centering
    \subfloat[American (ASL)]{
        \includegraphics[width=.5\linewidth]{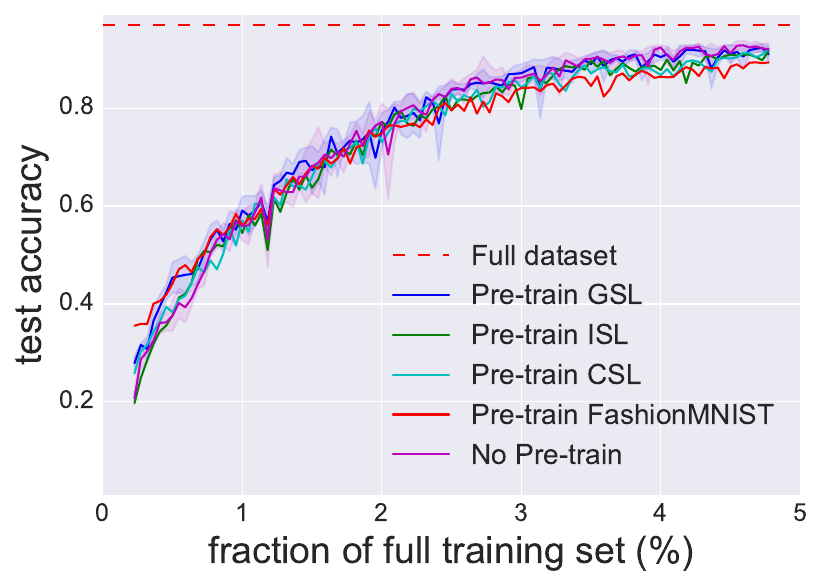}
    }
    \subfloat[Chinese (CSL)]{
	\includegraphics[width=.5\linewidth]{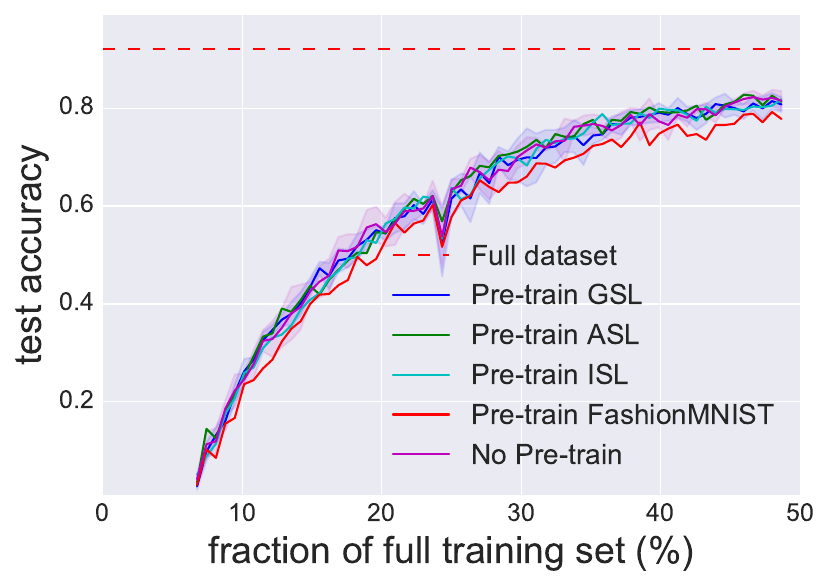}
    }
    \quad    
    \subfloat[German (GSL)]{
    	\includegraphics[width=.5\linewidth]{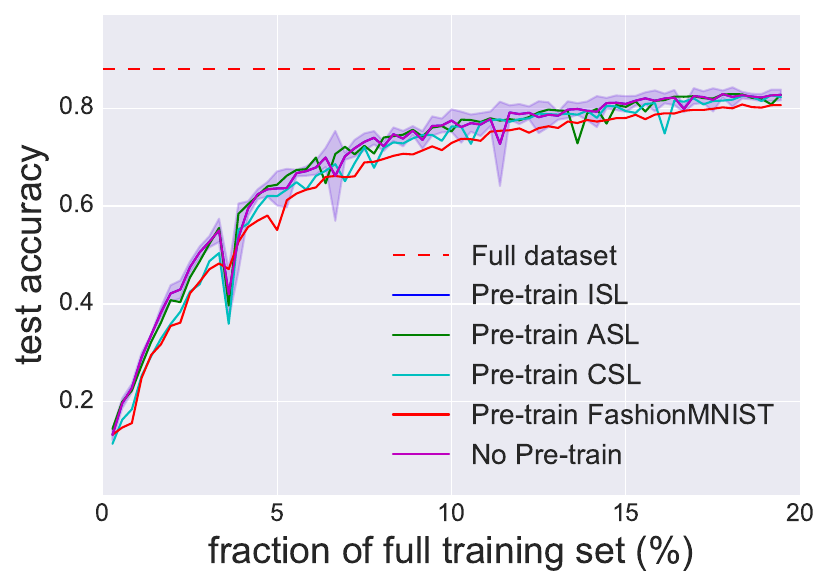}
    }
    \subfloat[Irish (ISL)]{
	\includegraphics[width=.5\linewidth]{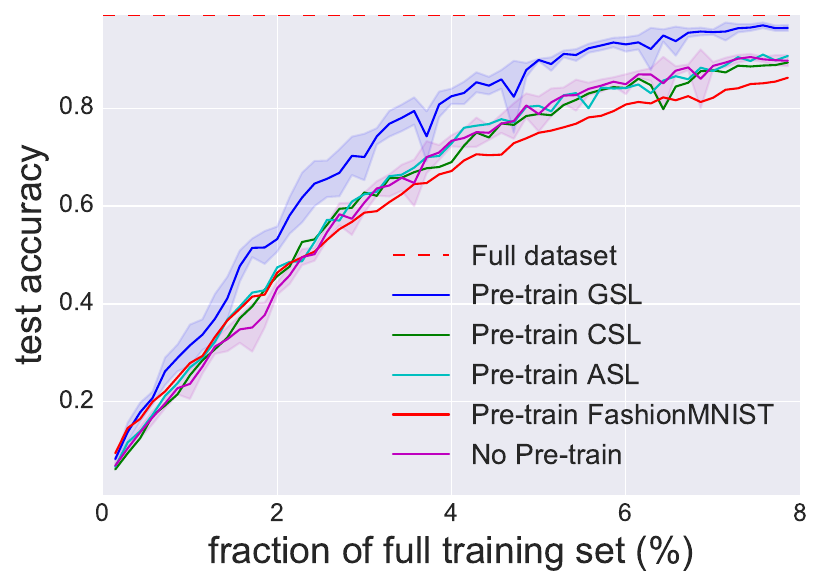}
    }
    \caption{\textit{Transfer Active Learning}: The figure above shows AL results for each fingerspelling corpus with pre-training and no pre-training.  Pre-training with a non-fingerspelling corpus degrades performance in all cases.}
    \label{fig:tran_act_res}
\end{figure}


\begin{figure}
    \centering
    \includegraphics[width=1\textwidth]{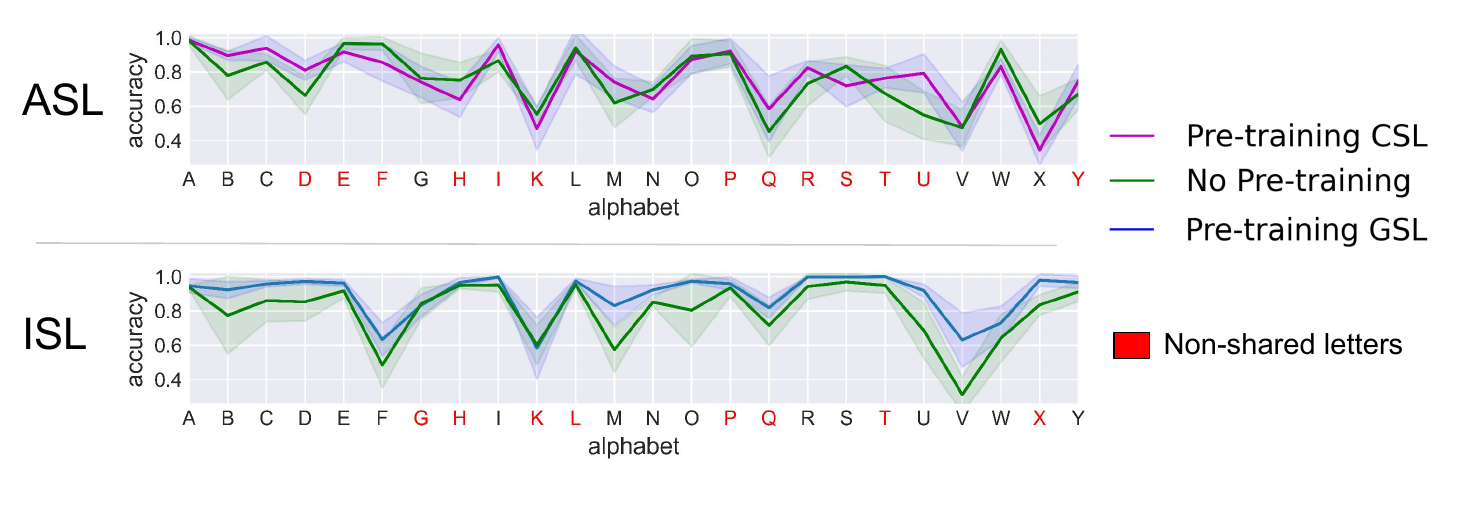}
    \vspace*{-10mm}
    \caption{\textit{Per Class Results for Transfer Active Learning}: Generally, we see the gains from pre-training are had by shared letters (red) but not exclusively (`X' in ISL is not shared by GSL).}
    \label{fig:con_mat}
\end{figure}

\subsection{Single Corpus Active Learning}

\paragraph{Implementation Details} We compare against uniformly random acquisition as the baseline. We re-run experiments with three unique random seeds to account for sampling variation.  The classifier is a neural network with two convolutional layers (ReLU activations and dropout with $p=25\%$) and two fully-connected layers (ReLU activations and dropout with $p=50\%$).

\paragraph{Results} Figure \ref{fig:active_res} reports the results, showing test set accuracy (y-axis) as more data is collected (x-axis).  We see a clear improvement over random sampling in all cases.  Moreover, AL reaches near full-data-set performance by seeing just $15\%$ (or less) of the original data in three out of four cases (ASL, GSL, and ISL).  GSL sees the least benefit from AL, but surprisingly, we see that with just $12\%$ of the training data, the model starts to surpass full-data-set performance.  We suspect that this is due to outliers (created by frame extraction and/or cropping mentioned in section \ref{Datasets}) that the AL method has not yet seen but detract from performance.

\subsection{Transfer Active Learning}
\paragraph{Implementation Details} We next turn to the pre-training experiment.  We use a ResNet-18 \citep{ResNet} as the backbone with the same structure as the previous experiment. We pre-train the classifier on the full data set from one corpus and then load the backbone of the pre-trained model with a re-initialized classification head.  The entire structure of the neural network is retrained during AL. We re-run all experiments five times with different random seeds. As a control, we also perform a run with the model pre-trained on FashionMNIST \citep{xiao2017fashionmnist}, another grayscale data set.  

\paragraph{Results}  Figure \ref{fig:tran_act_res} reports test accuracy for all pre-training configurations, including no pre-training.  The only noticeable trend that generalizes to all corpora is that pre-training on FashionMNIST degrades performance in comparison to pre-training on a fingerspelling corpus \emph{and} using no pre-training.  While this observation supports our motivating hypothesis for transfer AL, otherwise, we see modest gains with pre-training.  The only clear improvement over no pre-training is demonstrated for ISL pre-trained on GSL (subfigure d, blue line).  As these are the only two data sets that include the signer's forearm, we suspect that visual similarly is causing the benefit rather than linguistic relationship, or else we would have seen improvements in other data pairings. Yet, curiously, this benefit is not symmetric since pre-training with ISL did not improve GSL performance more than pre-training with other corpora.  Figure \ref{fig:con_mat} shows the accuracy per class (letter). To make the performance gap more obvious, we show performance at the time step with the largest gap ($t=40$).  We expect to see the pink and blue lines (pre-training) to be higher than the green line (no pre-training) for letters shown in black (shared across languages).  This is generally the case, but we also see improvements for non-shared letters, such as for `X' in ISL. 



\section{Conclusion}
We have reported the first transfer AL results for fingerspelling corpora.  We observed a clear success (pretraining on GSL and performing AL on ISL), but this was not observed for other pairings nor is the ISL-GSL relationship symmetric.  In future work, we plan to more carefully control for linguistic and visual relationships between data sets to better isolate the cause of the performance gains.


\vskip 0.2in
\section*{Acknowledgements} We thank Floris Roelofsen, Paul Verhagen, Rajeev Verma, and Haochen Wang for helpful feedback.
\bibliography{sample}

\newpage

\appendix



\section{Background}
\label{app:theorem}
\subsection{Sign language Recognition}
Hand gesture recognition is essential for applications such as human-computer
interaction, robot control, and sign language assistance systems~\citep{gesture_drones, silanon2014finger, sign_reco_overview}.
The purpose of sign language alphabet recognition is to detect labels from isolated or continuous Signs~\citep{yin-etal-2021-including}. Recognition of signs using computational models is a challenging problem because of a number of reasons. In some cases, one gesture can look very different from a different angle. Another challenge is the variations of how a sign is performed by different signers ~\citep{article_AUSTL}.
\cite{sign_recog} built ASL alphabet classifiers by a combination of 24 (only static) sign detectors with AdaBoost, then using threshold information given by each detector to suggest which gesture is the best match. \cite{act_sign} used active learning to utilize the availability of the current training classifier for increasingly improving performance. They do not use certain acquisition functions to measure uncertainty but directly choose the sample which is not correctly classified by the current classifier.

\subsection{Active learning and Transfer active learning}
Active learning is a machine learning method in which a learning system can query a human expert interactively to label data with the desired output\citep{AL_1996}. The main goal of active learning is that active learner performance is better than traditional supervised learning with the same labeled data. Therefore active learners can interactively pose queries during the training stage and are part of the human-in-the-loop paradigm.
Transfer Learning–Based Active Learning is a method to take advantage transfer learning for better active learning. Some papers tried to prove that pre-trained models are better active learners.  ~\citep{10.3389/frai.2021.635766} ~\citep{AL_Clinical_texts}

\subsection{Experiment for higher resolution of data}
\begin{figure}[htbp]
    \centering
    \subfloat[American (ASL)]{
        \includegraphics[width=.5\linewidth]{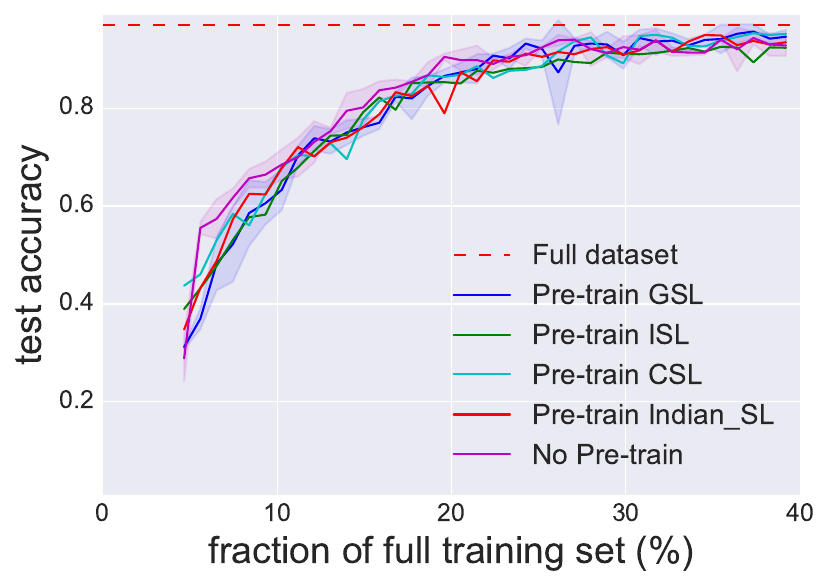}
    }
    \subfloat[Chinese (CSL)]{
	\includegraphics[width=.5\linewidth]{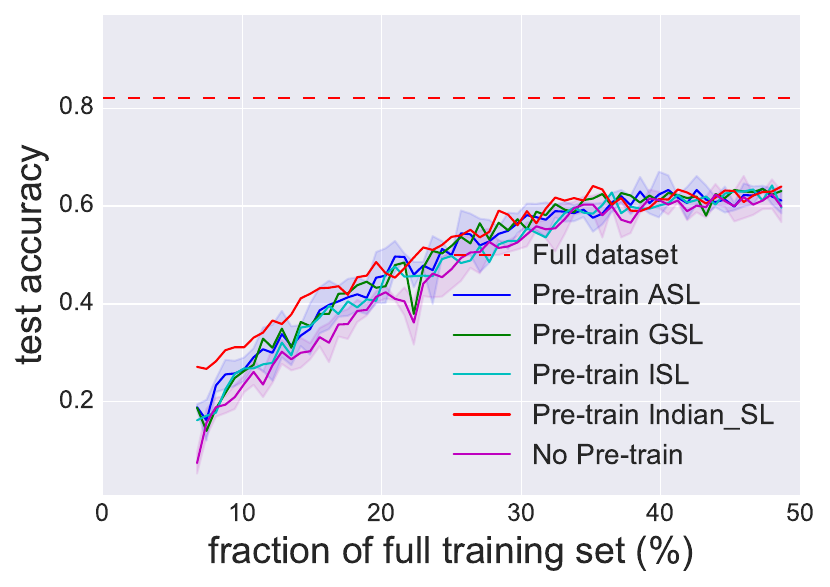}
    }
    \quad    
    \subfloat[German (GSL)]{
    	\includegraphics[width=.5\linewidth]{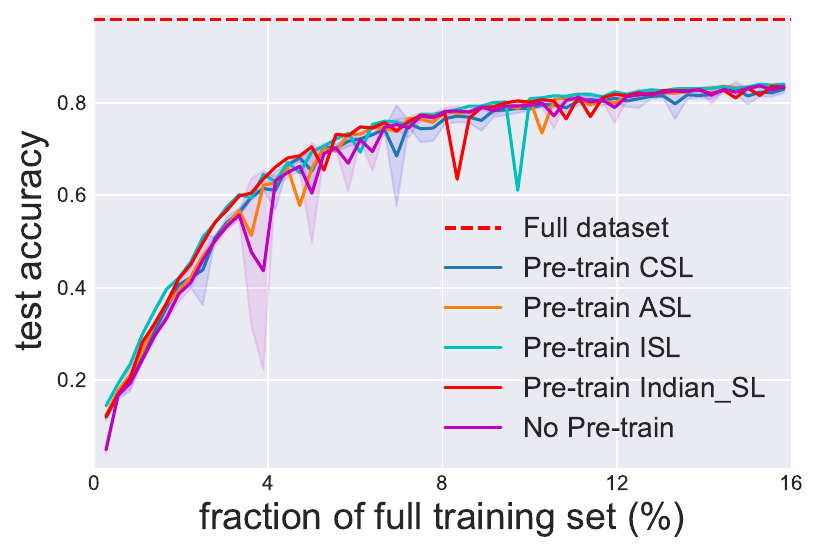}
    }
    \subfloat[Irish (ISL)]{
	\includegraphics[width=.5\linewidth]{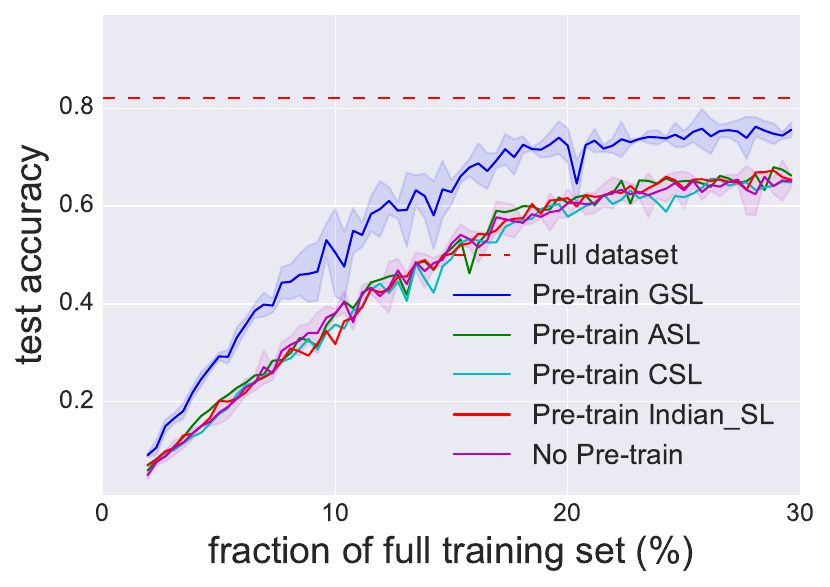}
    }
    \caption{\textit{Transfer Active Learning}: The figure above shows AL results for each fingerspelling corpus with pre-training and no pre-training for resolution = 96×96.  Pre-training with a double hand-fingerspelling(Indian\_SL) exists as a reference.}
    \label{fig:tran_act_res_96}
\end{figure}

\begin{figure}
    \centering
    \vspace*{-3mm}
    \includegraphics[width=1\textwidth]{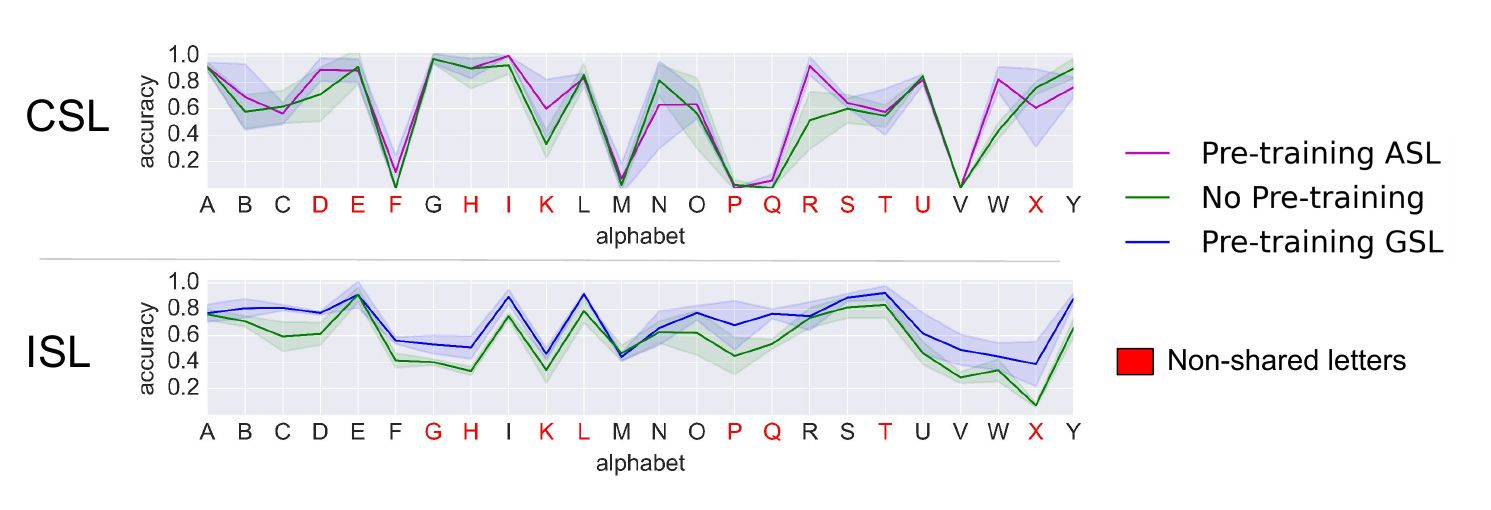}
    \vspace*{-7mm}
    \caption{\textit{Per Class Results for Transfer Active Learning at 96×96}: Generally, we see the gains from pre-training are had by shared letters (red) but not exclusively (`X' in ISL is not shared by GSL).}
    \label{fig:con_mat_96}
\end{figure}

 To ensure resolution does not limit the knowledge that the pre-trained model can transfer, we resize data in all four corpora to 96×96 and redo the same experiments. Because the ASL dataset used for the 28*28 experiment has a resolution of exactly 28*28, it cannot be used for resolution 96*96. We use a new ASL dataset from Kaggle\footnote{\hyperlink{https://www.kaggle.com/ayuraj/asl-dataset}{https://www.kaggle.com/ayuraj/asl-dataset}} with an original resolution of 400*400 for this experiment.
 Because the computation time grows rapidly with the resolution, we made some adjustments to save time. Firstly, we only repeated all experiments three times for transfer active learning. Secondly, because the Irish dataset is augmented with rotation, we sample $30\%$ of the Irish dataset and use it for experiments. Figure \ref{fig:tran_act_res_96} reports test accuracy for all pre-training configurations, including no pre-training and double-hand fingerspelling(Indian sign language). Similar to low-resolution results, the most obvious improvement is pre-train on GSL and fine-tuning on ISL. Figure \ref{fig:con_mat_96} shows the accuracy per class(gap). Interestingly, we also observe improvements for non-sharing alphabets, e.g. ’D’, ’Q’, ’R’ in ASL and ’P’, ’Q’, ’X’ in ISL. This shows that pre-training and fine-tuning can also help the model classify some non-shared alphabets.
 
\end{document}